\documentclass[conference]{IEEEtran}
\usepackage{cite}
\usepackage{amsmath,amssymb,amsfonts}
\usepackage{algorithmic}
\usepackage{graphicx}
\usepackage{textcomp}
\usepackage{xcolor}
\usepackage{fancyhdr}
\usepackage[hyphens]{url}

\usepackage{multirow,makecell}
\usepackage[para,online,flushleft]{threeparttable}
\usepackage[switch]{lineno}
\usepackage[caption=false]{subfig}
\usepackage{multicol}
\usepackage{booktabs} % For formal tables
\usepackage{comment}

\newcommand{\hw}[1]{\ensuremath{\mathtt{#1}}}

\def\BibTeX{{\rm B\kern-.05em{\sc i\kern-.025em b}\kern-.08em
    T\kern-.1667em\lower.7ex\hbox{E}\kern-.125emX}}

\title{ILMPQ : An Intra-Layer Multi-Precision Deep Neural Network Quantization framework for FPGA } 
\author{\IEEEauthorblockN{Sung-En Chang$^{*1}$, Yanyu Li$^{*1}$, Mengshu Sun$^{*1}$, \thanks{$^*$Equal contribution. }Yanzhi Wang$^1$,
Xue Lin$^1$}
\IEEEauthorblockA{\textit{$^1$Northeastern University, } \\
\{chang.sun, li.yanyu, sun.meng, yanz.wang, xue.lin\}@northeastern.edu}
}
\begin{document}
\maketitle
\pagestyle{plain}

%%%%%% -- PAPER CONTENT STARTS-- %%%%%%%%

\begin{abstract}

This work targets the commonly used FPGA (field-programmable gate array) devices as the hardware platform for DNN edge computing. 
We focus on DNN quantization as the main model compression technique. %, since DNN quantization has great importance in terms of reducing model size and computation cost on varieties of hardware platforms. 
The novelty of this work is: We use a quantization method that supports multiple precisions along the intra-layer dimension, while the existing quantization methods apply multi-precision quantization along the inter-layer dimension. The intra-layer multi-precision method can uniform the hardware configurations for different layers to reduce computation overhead and at the same time preserve the model accuracy as the inter-layer approach. Our proposed ILMPQ DNN quantization framework achieves $70.73\%$ Top1 accuracy in ResNet-18 on the ImageNet dataset. We also validate the proposed MSP framework on two FPGA devices i.e., Xilinx XC7Z020 and XC7Z045.
We achieve $3.65\times$ speedup in end-to-end inference time on the ImageNet, comparing with the fixed-point quantization method.
\end{abstract}
\section{Introduction}

Deep neural net (DNN) quantization is an crucial technique to reduce the computation, memory, and storage requirements executing on-device inference, especially for platforms with capability of customized architecture design, such as FPGA devices and ASIC chips.

Various quantization schemes have been proposed, such as binary~\cite{courbariaux2015binaryconnect},
ternary\cite{li2016ternary}, and Power-of-Two (PoT)~\cite{DBLP:journals/corr/MiyashitaLM16}. These works can significantly reduce the computation but suffer from non-neglectable accuracy degradation. On the other hand,
Fixed-point (Fixed)~\cite{zhou2016dorefa,choi2018pact} 
yields relatively small accuracy loss while still needs expensive multiplication operations during inference computation.

\begin{figure}[t]
\centering  
\includegraphics[width=1.0\columnwidth]{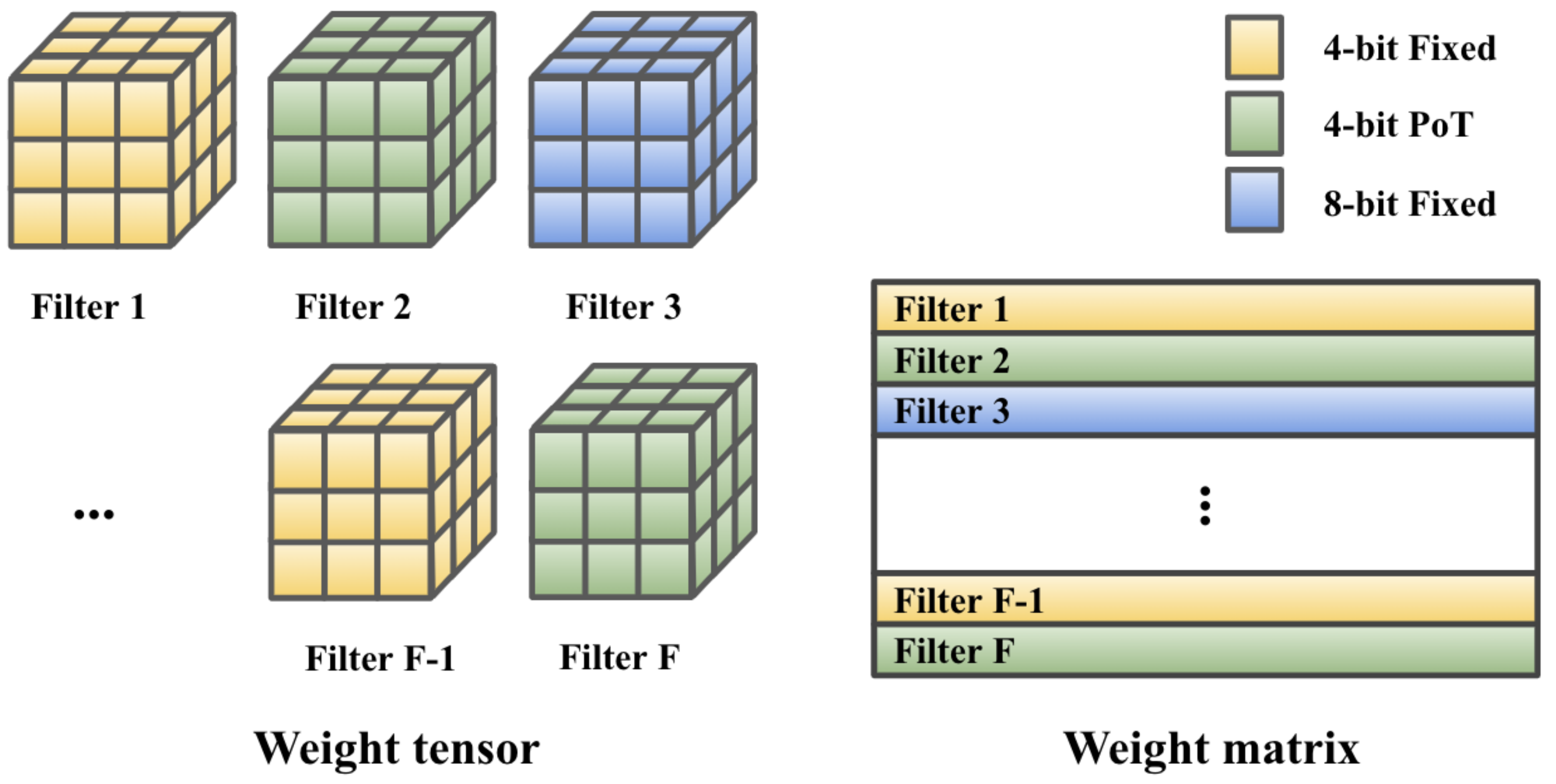}
\caption{\textbf{The proposed DNN quantization framework with filter-wise mixed schemes and multiple precision, which assigns quantization precision and scheme to filters of the weight tensor (or rows of the weight matrix).} 
}
\label{fig:IntraLayer}
\end{figure}

To address this issue, this work proposes a novel DNN quantization framework, namely ILMPQ, which is Intra-Layer Multi-precision quantization approach.
%Figure \ref{fig:IntraLayer} illustrates the proposed quantization framework on the DNN (a) weight tensor and (b) weight matrix.
Specifically, Figure~\ref{fig:IntraLayer} shows that each filter of the weight tensor or each row in the weight matrix can be assigned with a specific configuration of quantization scheme and precision. 
The candidates of schemes and precisions are %derived \emph{practically} 
assigned to facilitate most efficient hardware implementation, while with the capability to preserve the accuracy as the unquantized (32-bit floating-point) baseline models.
This highly hardware-informative quantization strategy significantly reduces the search space of the DNN quantization problem, making our framework distinctive from existing multi-precision quantization work. 
 
The contributions of our quantization framework are summarized as follows:
\begin{itemize}
    
    \item \textbf{We propose an \emph{Intra-Layer Flexibility}, which can be applied to all layers in a DNN model, achieving the best accuracy performance without damaging hardware efficiency.}
    
    \item \textbf{A hardware-aware solution, which can significantly reducing the problem search space.}
    
    \item \textbf{The significant inference speedup on real devices, comparing to the network-wise uniform low-bit quantization i.e., the speed upper bound of existing layer-wise multi-precision approaches.}

\end{itemize}

\section{Proposed ILMPQ Quantization}

% new
\begin{table*}[t]
\small
\centering
\tabcolsep 2.2pt
\begin{tabular}{c|c|c||c|c||c|c|c|c||c|c|c|c}
\toprule
\multirow{3}{*}{\makecell{Quantization \\ Method}} & \multirow{3}{*}{\makecell{PoT-4 \\: Fixed-4 \\: Fixed-8}} & \multirow{3}{*}{\makecell{First/Last \\ Layer \\ Quantization}} & \multicolumn{2}{c||}{Accuracy} & \multicolumn{4}{c||}{Results on FPGA XC7Z020} & \multicolumn{4}{c}{Results on FPGA XC7Z045} \\
\cline{4-13}
 & & & Top-1 & Top-5 & \multicolumn{2}{c|}{Utilization} & Throughput & Latency & \multicolumn{2}{c|}{Utilization} & Throughput & Latency \\
\cline{6-7} \cline{10-11}
 & & & (\%) & (\%) & LUT & DSP & (GOP/s) & (ms) & LUT & DSP & (GOP/s) & (ms) \\
\hline
(1) Fixed & 0:100:0 & 8-bit Fixed & 69.72 & 88.67 & 49\% & 100\% & 29.6 & 122.6 & 21\% & 100\% & 115.6 & 31.4 \\
(2) Fixed & 0:100:0 & \checkmark & 68.66 & 87.54 & 45\% & 100\% & 36.5 & 99.3 & 24\% & 100\% & 142.7 & 25.4 \\
(3) PoT & 100:0:0 & 8-bit Fixed & 68.20 & 87.14 & 51\% & 100\% & 62.4 & 58.1 & 40\% & 100\% & 290.5 & 12.5 \\
(4) PoT & 100:0:0 & \checkmark & 67.11 & 85.93 & 57\% & 12\% & 72.2 & 50.2 & 44\% & 3\% & 352.6 & 10.3 \\
(5) PoT + Fixed & 50:50:0 & 8-bit Fixed & 68.94 & 88.66 & 71\% & 100\% & 50.3 & 72.0 & 42\% & 100\% & 196.8 & 18.4 \\
(6) PoT + Fixed & 50:50:0 & \checkmark & 67.98 & 86.75 & 66\% & 100\% & 75.8 & 47.8 & 38\% & 100\% & 296.3 & 12.2 \\
(7) PoT + Fixed & 60:40:0 & 8-bit Fixed & 68.53 & 88.47 & 80\% & 100\% & 57.0 & 63.6 & - & - & - & - \\
(8) PoT + Fixed & 67:33:0 & 8-bit Fixed & 68.46 & 88.22 & - & - & - & - & 61\% & 100\% & 245.8 & 14.8 \\
\bf{ILMPQ-1} & 60:35:5 & \checkmark & \bf{70.66} & \bf{89.53} & 82\% & 100\% & \bf{89.0} & \bf{40.7} & - & - & - & - \\
\bf{ILMPQ-2} & 65:30:5 & \checkmark & \bf{70.73} & \bf{89.62} & - & - & - & - & 65\% & 100\% & \bf{421.1} & \bf{8.6} \\
\bottomrule
\end{tabular}
\caption{\textbf{Implementations of different quantization schemes on two FPGA boards (XC7Z020 and XC7Z045) for ResNet-18 on ImageNet, using the (equivalent) 4-bit precision.} 
}
\label{tab:ablation}
\end{table*}

\subsection{Intra-Layer flexibility to preserve the accuracy}

We propose \emph{Intra-Layer Multi-precision quantization} for our quantization scheme. In each layer, we only quantize 5 percent filters of weights to 8 bit and leave the rest to 4 bit. Rather than the prior works which need to use 8 or more bits to represent the weights in the first and the last layer. Our ILMPQ can be applied in the all layers of the DNN models. This is because in \emph{intra-layer flexibility}, the weights quantized using 8 bits can be trained to mitigate the imprecision caused by those weights quantized using fewer bits. This mitigation happens in every layer. %On the other hand, in the prior inter-layer flexibility, the majority of layers will be quantized with fewer bits. The resultant imprecision cannot be mitigated within a layer and will be accumulated across layers. It is difficult to recover the accumulated imprecision by limited layers quantized with more bits.

Besides algorithm-level advantages, the proposed intra-layer flexibility also exhibits an advantage at the FPGA hardware level. Recall that the same quantization scheme (e.g., 4-bit for $95\%$ of weights and 8-bit for the rest of $5\%$) is applied to all layers of a DNN. At FPGA configuration time for a specific DNN inference task, one could allocate a portion of PEs for the low-bit portion of computation and the rest of PEs for the 8-bit portion, and this works for every layer. As for traditional inter-layer multi-precision scheme, it's almost impossible to perform online reconfiguration, that is, the PEs assigned to execute 8-bit first/last layers is vacant while processing the middle layers.

\subsection{Mixed Schemes to boost the hardware efficiency}

Inspired by MSQ~\cite{chang2020mix}, which is the state-of-the-art, FPGA specific quantization scheme. We incorporate their Mixed-Scheme quantization to our work to further boost the hardware efficiency.

As inference on FPGA is conducted layer-by-layer on GEMM core, different schemes within a layer can benefit hardware parallelism. Specifically, fixed-point convolution is done by \hw{GEMM_{Fixed}} and PoT is done by \hw{GEMM_{PoT}}. 
We implement \hw{GEMM_{Fixed}} on DSP modules and \hw{GEMM_{PoT}} on LUT modules on FPGA, so that the heterogeneous resource can be utilized efficiently. As different FPGA device has different characteristics, the actual mixing ratio of fixed-point scheme and PoT scheme can be determined offline by examining FPGA throughput. The ideal utilization is to balance workload of the two schemes and achieve highest throughput. 

\subsection{The training process of ILMPQ framework}
 
In the training algorithm of our ILMPQ quantization, there are two steps to determine the bitwidth and schemes within each layer. First, we compute the largest eigenvalue of Hessian matrix per filter to determine the most sensitive weights. More bits will be assigned to filters with larger eigenvalues. 
Then, we sort the row vectors by their variance. Rows with smaller variance are quantized to PoT, while others are quantized to fixed-point scheme. The ratio is determined offline by hardware utilization. 
The reason behind this assignment is that PoT scheme enjoys higher resolution around the mean area (zero) compared to fixed-point scheme, so that quantization error can be reduced if the weights to be quantized mostly fall around zero, which empirically means lower variance of weight distribution.

\section{Experiments and Evaluation}\label{sec:eva}

To present the accuracy and hardware performance with real-world applications. 
We compare ILMPQ with other quantization methods on the ImageNet dataset using ResNet-18, as displayed in Table~\ref{tab:ablation}.
The quantized model is trained under basic data augmentation and step learning rate on PyTorch platform. Initialized with pretrained model, the quantization-aware training takes 50 epochs.
We provide hardware results on two FPGA boards, XC7Z020 and XC7Z045. 
Specifically, the optimal ratio on XC7Z020 is 60:35:5 (ILMPQ-1), resulting in top-1 accuracy of 70.66\% and latency of 40.7ms, while the optimal ratio on XC7Z045 is 65:30:5 (ILMPQ-2), leading to top-1 accuracy of 70.73\% and latency of 8.6ms.
ILMPQ with the optimal ratio of quantization schemes achieves up to $3.01\times$ speedup on XC7Z020 and up to $3.65\times$ speedup on XC7Z045, with both LUTs and DSPs utilized efficiently.

\section{Conclusion}\label{sec:con}
In this work, we propose a new dimension of DNN quantization, namely, ILMPQ, which introduce the strategy of row-wise mixed scheme and intra-layer mixed precision. We achieve state-of-art accuracy performance and up to $3.65\times$ speed up on FPGA.

\section*{Acknowledgment}

This work is partly supported by the National Science Foundation CCF-1901378, CCF-1919117, and CCF-1937500.

%\section*{Acknowledgements}
%This document is derived from previous conferences, in particular ISCA 2020.

%%%%%%% -- PAPER CONTENT ENDS -- %%%%%%%%

%%%%%%%%% -- BIB STYLE AND FILE -- %%%%%%%%
\bibliographystyle{IEEEtranS}
\bibliography{refs}
%%%%%%%%%%%%%%%%%%%%%%%%%%%%%%%%%%%%

\end{document}